\pdfoutput=1
\documentclass[11pt]{article}

\usepackage[]{acl}

\usepackage{times}
\usepackage{latexsym}

\usepackage[T1]{fontenc}
\usepackage[utf8]{inputenc}

\usepackage{algorithm}
\usepackage{microtype}
\usepackage{algcompatible,amsmath}
\usepackage{graphicx}
\usepackage{url}
\usepackage{multirow}
\usepackage{comment}
\usepackage{amssymb}
\usepackage{pifont}
\usepackage{balance}
\usepackage{mathtools}
\usepackage{xcolor}
\usepackage{colortbl}
\usepackage{bbm}
\usepackage{enumitem}
\usepackage{marvosym}
\usepackage{textcomp}
\usepackage{array}
\usepackage{booktabs}
\usepackage{tabularx}
\usepackage{CJKutf8}

\usepackage{newfloat}
\usepackage{listings}
\floatstyle{ruled}
\newfloat{listing}{tb}{lst}{}
\floatname{listing}{Listing}

\newcommand{\squishlist}{
\begin{list}{$\bullet$}
{ \usecounter{Lcount}
\setlength{\itemsep}{0pt}
\setlength{\parsep}{0pt}
\setlength{\topsep}{0pt}
\setlength{\partopsep}{0pt}
\setlength{\leftmargin}{2em}
\setlength{\labelwidth}{1.5em}
\setlength{\labelsep}{0.5em} } }
\newcommand{\squishend}{
\end{list} }

\newcommand{\cmark}{\ding{51}}%
\newcommand{\xmark}{\ding{55}}%

\title{CHEF: A Pilot Chinese Dataset for Evidence-Based Fact-Checking}

\author{Xuming Hu$^{1*}$, Zhijiang Guo$^{2*}$, Guanyu Wu$^1$, Aiwei Liu$^1$, Lijie Wen$^{1\dagger}$, Philip S. Yu$^{1,3}$\\
  $^1$Tsinghua University\\  $^2$University of Cambridge\\ $^3$University of Illinois at Chicago\\
  $^1$\texttt{\{hxm19,wugy18,liuaw20\}@mails.tsinghua.edu.cn}\\
  $^2$\texttt{zg283@cam.ac.uk}
  $^1$\texttt{wenlj@tsinghua.edu.cn}
  $^3$\texttt{psyu@uic.edu}\\
  }

\begin{document}
\maketitle
\begin{abstract}

The explosion of misinformation spreading in the media ecosystem urges for automated fact-checking. While misinformation spans both geographic and linguistic boundaries, most work in the field has focused on English. Datasets and tools available in other languages, such as Chinese, are limited. In order to bridge this gap, we construct CHEF, the first CHinese Evidence-based Fact-checking dataset of 10K real-world claims. The dataset covers multiple domains, ranging from politics to public health, and provides annotated evidence retrieved from the Internet. Further, we develop established baselines and a novel approach that is able to model the evidence retrieval as a latent variable, allowing jointly training with the veracity prediction model in an end-to-end fashion. Extensive experiments show that CHEF will provide a challenging testbed for the development of fact-checking systems designed to retrieve and reason over non-English claims. Source code and data are available\footnote{\url{https://github.com/THU-BPM/CHEF}\\\phantom{00} $^{*}$Equally Contributed.\\\phantom{00} $^\dagger$Corresponding Author.}.

\end{abstract}
\section{Introduction}
\label{sec:intro}

Misinformation is being spread online at increasing rates, posing a challenge to media platforms from newswire to social media. In order to combat the proliferation of misinformation, fact-checking is an essential task that assesses the veracity of a given claim based on evidence~\citep{Vlachos2014FactCT}. Fact-checking is commonly conducted by journalists.
However, fact-checking is a time-consuming task, which can take journalists several hours or days~\citep{Adair2017ProgressT}. Thus, there is a need for automating the process.

\begin{CJK*}{UTF8}{gbsn}
\begin{table}[t]
\centering
\scalebox{0.85}{
\begin{tabularx}{9cm}{X}
\toprule
\textbf{Claim}: 2019年, 共有12.08万人参加成都中考，但招生计划只有4.3万。\textit{In 2019, a total of 120,800 students participated in the high school entrance examination in Chengdu, but schools only enrolled 43,000 students.} \\
\midrule
\textbf{Document}: \textcolor{blue}{今年共有12.08万人参加中考，这个是成都全市, 包括了20个区，高新区和天府新区的总参考人数。} 月前，教育局公布了2019年的普高招生计划。招生计划数进一步增加，上普高的机会更大了... \textcolor{blue}{中心城区（13个区）招生计划为43015人。} \textit{\textcolor{blue}{This year, 120,800 people participated in the high school entrance examination. This number is for the entire city of Chengdu, including 20 districts, high-tech zone and Tianfu new district.} A month ago, the Education Bureau announced the 2019 high school enrollment plan. The number of enrollment will be increased, indicating that there is a greater chance of going to high school... \textcolor{blue}{The plan of the central area (including 13 districts) is 43,015.}} \\
\midrule
\textbf{Verdict}: Refuted; \textbf{Domain}: Society \\
\midrule
\textbf{Challenges}: Evidence Collection; Numerical Reasoning \\
\bottomrule
\end{tabularx}}
\vspace{-2mm}
\caption{An example from CHEF (Chinese is translated into English). The claim is refuted by the evidence, which are sentences retrieved (\textcolor{blue}{highlighted}) from the document. For brevity, only the relevant snippet of the document is shown. }
\label{tab:example}
\vspace{-3mm}
\end{table}
\end{CJK*}

\begin{table*}
\centering
\scalebox{0.7}{
\begin{tabular}{lcccccccc}
\toprule
\multirow{2}{*}{Dataset} & \multirow{2}{*}{Natural} & \multirow{2}{*}{Domain} & \multirow{2}{*}{\#Claims} & \multirow{2}{*}{Language}
& \multicolumn{4}{c}{Evidence} \\ 
\cline{6-9}
& & & & & Type & Source &  Retrieved & Annotated  \\\midrule
FEVER~\citep{Thorne2018FEVERAL} & \textcolor{red}{\xmark} & Multiple & 185,445 & English & Text & Wiki & \textcolor{blue}{\cmark} & \textcolor{blue}{\cmark} \\
HOVER~\citep{jiang2020hover} & \textcolor{red}{\xmark} & Multiple & 26,171  & English & Text & Wiki & \textcolor{blue}{\cmark} & \textcolor{blue}{\cmark} \\
TabFact~\citep{Chen2020TabFactAL} & \textcolor{red}{\xmark} &  Multiple & 92,283  & English & Table & Wiki & \textcolor{red}{\xmark} & \textcolor{blue}{\cmark} \\
InfoTabs~\citep{Gupta2020INFOTABSIO} & \textcolor{red}{\xmark} &  Multiple & 23,738 & English & Table & Wiki & \textcolor{red}{\xmark} & \textcolor{blue}{\cmark} \\
ANT~\citep{Khouja2020StancePA} & \textcolor{red}{\xmark} &  Multiple & 4,547 & Arabic  & \textcolor{red}{\xmark} & \textcolor{red}{\xmark} & \textcolor{red}{\xmark} & \textcolor{red}{\xmark} \\
VitaminC~\citep{vitaminc2021} & \textcolor{red}{\xmark} &  Multiple & 488,904 & English & Text & Wiki & \textcolor{red}{\xmark} & \textcolor{blue}{\cmark} \\
DanFEVER~\citep{danfever2021} & \textcolor{red}{\xmark} &  Multiple & 6,407 & Danish & Text & Wiki & \textcolor{blue}{\cmark} & \textcolor{blue}{\cmark} \\
FEVEROUS~\citep{Aly2021FEVEROUSFE} & \textcolor{red}{\xmark} &  Multiple & 87,026  & English  & Text/Table & Wiki & \textcolor{blue}{\cmark} & \textcolor{blue}{\cmark} \\
\midrule
PolitiFact~\citep{Vlachos2014FactCT} & \textcolor{blue}{\cmark} & Politics & 106 & English & Meta/Text & FC & \textcolor{red}{\xmark} & \textcolor{red}{\xmark} \\
PunditFact~\citep{Rashkin2017TruthOV} & \textcolor{blue}{\cmark} & Multiple & 4,361 & English & \textcolor{red}{\xmark} & \textcolor{red}{\xmark} & \textcolor{red}{\xmark} & \textcolor{red}{\xmark} \\
Liar~\citep{Wang2017LiarLP} & \textcolor{blue}{\cmark} & Multiple & 12,836  & English & Meta & FC & \textcolor{red}{\xmark} & \textcolor{red}{\xmark} \\
Verify~\citep{Baly2018IntegratingSD} &  \textcolor{blue}{\cmark} & Politics & 422  & Mul(2) & Text & Internet  & \textcolor{blue}{\cmark} & \textcolor{red}{\xmark} \\
MultiFC~\citep{augenstein2019multifc} &  \textcolor{blue}{\cmark} & Multiple & 36,534  & English & Meta/Text & Internet & \textcolor{blue}{\cmark} & \textcolor{red}{\xmark} \\
Snopes~\citep{Hanselowski2019ARA} &  \textcolor{blue}{\cmark} & Multiple & 6,422  & English  & Text & FC & \textcolor{red}{\xmark} & \textcolor{red}{\xmark} \\
% WikiFactCheck~\citep{Gupta2020INFOTABSIO} & \textcolor{blue}{\cmark} &  Multiple & 124,841 & English & Text & Wiki & \textcolor{red}{\xmark} & \textcolor{blue}{\cmark} \\
SciFact~\citep{Wadden2020FactOF} & \textcolor{blue}{\cmark} & Science & 1,409   & English & Text & Paper & \textcolor{red}{\xmark} & \textcolor{red}{\xmark} \\
PUBHEALTH~\citep{kotonya2020explainable} & \textcolor{blue}{\cmark} & Health & 11,832   & English & Text & FC & \textcolor{red}{\xmark} & \textcolor{red}{\xmark} \\
AnswerFact~\citep{Zhang2020AnswerFactFC} & \textcolor{blue}{\cmark} & Product & 60,864   & English  & Meta/Text & Amazon & \textcolor{blue}{\cmark} & \textcolor{red}{\xmark} \\
FakeCovid~\citep{shahifakecovid} &  \textcolor{blue}{\cmark} & Health & 5,182  & Mul(3)  & \textcolor{red}{\xmark} & \textcolor{red}{\xmark} & \textcolor{red}{\xmark} & \textcolor{red}{\xmark} \\
XFact~\citep{gupta2021} &  \textcolor{blue}{\cmark} & Multiple & 31,189  & Mul(25)  & Meta/Text & Internet & \textcolor{blue}{\cmark} & \textcolor{red}{\xmark} \\
\midrule
CHEF & \textcolor{blue}{\cmark} & Multiple & 10,000 & Chinese & Meta/Text & Internet & \textcolor{blue}{\cmark} & \textcolor{blue}{\cmark} \\
\bottomrule
\end{tabular}}
\vspace{-2mm}
\caption{Comparisons of fact-checking datasets. Type in the header means the type of evidence used, which can be text, metadata or both. Source means where the evidence are collected from, such as Wikipedia (Wiki), fact-checking websites (FC). Retrieved denotes if the evidence is given or retrieved from the source. Annotated means whether the evidence is manually annotated. Verify, FakeCovid, XFact contain claims in multiple languages, but Chinese claims are not included.}
\label{tab:dataset}
\vspace{-2mm}
\end{table*}

Although misinformation spans both geographic and linguistic boundaries, most existing works focused on English~\citep{Wang2017LiarLP,Thorne2018FEVERAL,augenstein2019multifc,Hanselowski2019ARA, Chen2020TabFactAL}. There only exists a handful of non-English datasets for verifying real-world claims. However, these datasets are either small in size~\citep{Baly2018IntegratingSD}, or designed for multilingual systems~\citep{gupta2021}. On the other hand, \citet{Khouja2020StancePA} and \citet{danfever2021} created claims by paraphrasing sentences from non-English articles, but synthetic claims cannot replace real-world claims for training generally applicable fact-checking systems.

To bridge this gap, we introduce a CHinese dataset for Evidence-based Fact-checking (CHEF). CHEF includes claims that are not only relevant to the Chinese world, but also originally made in Chinese. It consists of 10,000 real-world claims, collected from 6 Chinese fact-checking websites covered multiple domains and paired with annotated evidence. To ensure annotation consistency, we developed suitable guidelines and performed data validation\footnote{The annotation guideline is provided in the appendix.}. We shared some of the insights obtained during the annotation process that we hope will be beneficial to other non-English annotation efforts. Table~\ref{tab:example} shows an instance from CHEF. In order to verify the claim, one needs to first retrieve the evidence sentences from related documents (e.g.\ government reports), then predict the veracity based on the evidence. After comparing the statistics of the entire city and central area, we can reach the verdict that the claim is refuted by evidence. To characterize the challenge of the dataset presented, we perform a thorough analysis and demonstrate the utility of the dataset by developing two types of baselines, including pipeline and joint systems. Our key contributions are summarized as follows:

% \squishlist
\begin{enumerate}
    \item We provide the first sizable multi-domain Chinese dataset for automated fact-checking. It consists of 10K real-world claims with manually annotated evidence sentences.
    \item We further propose an approach that is able to model the evidence selection as a latent variable, which can be jointly trained with the veracity prediction module.  
    \item We develop several established baselines and conduct a detailed analysis of the systems evaluated on the dataset, identifying challenges that need to be addressed in future research.
% \squishend
\end{enumerate}

\section{Background: Dataset Comparisons}
\label{sec:related}

In this section, we reviewed the existing fact-checking dataset as summarized in Table~\ref{tab:dataset}. Following~\citet{guo2022}, we grouped the datasets into two categories: natural and synthetic. Natural datasets consist of real-world claims, while synthetic datasets contain claims created artificially by mutating sentences from Wikipedia articles.

\subsection{Non-English Dataset}
Existing efforts in the construction of non-English datasets are limited, both in scope and in size. Verify~\citep{Baly2018IntegratingSD} contains 422 claims in Arabic and FakeCovid~\citep{shahifakecovid} has 3,306 non-English claims about COVID-19. Though XFact~\citep{Gupta2020INFOTABSIO} includes 31,189 claims in 25 languages, it mainly focuses on the multilingual setting, where the average number of instances per language is 1,248. More importantly, these datasets do not include annotated evidence. For example, XFact used search summaries returned by Google as evidence. On the other hand, \citet{Khouja2020StancePA} and \citet{danfever2021} constructed synthetic datasets by mutating sentences from Arabic news and Danish Wikipedia articles, respectively. Unlike previous efforts, CHEF consists of 10K real-world claims paired with annotated evidence.

There exist Chinese datasets focused on rumor detection~\citep{Ma2016DetectingRF, zhang2021mining}, which is classified into claim detection~\citep{kotonya20survey,guo2022}, as it is based on language subjectivity and growth of readership~\citep{Qazvinian2011RumorHI}. A claim can be factual regardless of whether it is a rumour~\citep{Zubiaga2018DetectionAR}. Unlike existing rumor detection datasets, CHEF focuses on factuality of the claim.

\subsection{Evidence-Based Fact-Checking}
Early efforts predicted the veracity solely based on the claims or with metadata~\citep{Rashkin2017TruthOV, Wang2017LiarLP}, but relying on surface patterns of claims without considering the state of the world fails to identify well-presented misinformation~\citep{Schuster2019TheLO}. Therefore, synthetic datasets~\citep{Thorne2018FEVERAL,jiang2020hover,Aly2021FEVEROUSFE} considered Wikipedia as the source of evidence and annotated the sentences supporting or refuting each claim. However, these efforts restricted world knowledge to a single source (i.e. Wikipedia), ignoring the challenge of retrieving evidence from heterogeneous sources on the Internet. 

To address this, recent natural datasets~\citep{augenstein2019multifc,gupta2021} used the summary snippets returned by Google as evidence. One key limitation of this approach is that summary snippets do not provide sufficient information to verify the claim. \citet{gupta2021} showed that only 45\% of snippets provide sufficient information, while 83\% of the full text from web pages provides sufficient evidence to determine veracity of the claim. To construct a better evidence-based dataset, we retrieve documents from web pages and manually select relevant evidence sentences from documents as evidence. Such a design makes CHEF suitable to train fact-checking systems that can extract evidence from web-sources and validate real-world claims based on evidence found on the Internet.

\section{Dataset Construction}
\label{sec:construction}

CHEF is constructed in four stages: data collection, claim labeling, evidence retrieval and data validation. Data collection selects sources, crawls claims and associated metadata. Claim labeling identifies claims from fact-checking articles and assigns the veracity labels of claims based on the article. Evidence retrieval collects documents from the Internet and selects the most relevant sentences as evidence. Data validation controls the annotation quality. The annotation team has 25 members, 5 of them are only involved in data validation. All annotators are native Chinese speakers. To ensure the annotation quality, they were trained by the authors and went through several pilot annotations.  

\begin{table}[t]
\centering
\scalebox{0.8}{
\begin{tabular}{lccc}
\toprule
Website & Domain & URL & Total \\
\midrule
Piyao & Multiple & \url{www.piyao.org.cn} & 3,741 \\
TFC & Multiple & \url{tfc-taiwan.org.tw} & 1,759 \\
Mygopen & Multiple & \url{www.mygopen.com} & 1,654 \\
Jiaozhen & Multiple & \url{vp.fact.qq.com} & 157 \\
\midrule
Cnews & Multiple & \url{m.chinanews.com} & 2,689 \\
\midrule
Total & Multiple & - & 10,000 \\
\bottomrule
\end{tabular}}
\vspace{-2mm}
\caption{Statistics of data sources. Piyao, TFC, MyGoPen and Jiaozhen are fact-checking websites. Cnews is a news website. }
\label{tab:stats}
% \vspace{-5mm}
\end{table}

\subsection{Data Collection}
We crawled all active Chinese fact-checking websites listed by Duke Reporters\footnote{\url{www.reporterslab.org/fact-checking/}}. However, most claims fact-checked by the fact-checkers are non-factual, solely relying on such claims will lead to an imbalance dataset. Therefore, we followed~\citet{kotonya2020explainable} by crawling articles from the news review site. As shown in Table~\ref{tab:stats}, this resulted in 5 websites in total. From each website, we crawled the full text of the article and corresponding metadata (e.g. author, domain, URL publication date). Totally, we crawled 14,770 fact-checking and news articles. There exists a number of crawling issues, such as the article could not be retrieved, or the content is not textual. We removed such instances. Next, we checked the dataset for duplications. Upon manual inspections, this was mainly due to them appearing on different websites. All duplications would be in the training split of the dataset, so that the model would not have an unfair advantage.  As shown in Figure~\ref{fig:domain}, claims cover multiple domains, including politics, public health, science, society and culture. More than 36\% of claims belong to public health domain, as many fact-checking articles focused on countering misinformation related to COVID-19. The society domain has second most claims, which involves social events that are closely related to people's daily lives.

\begin{figure}
    \centering
    \includegraphics[scale=0.35]{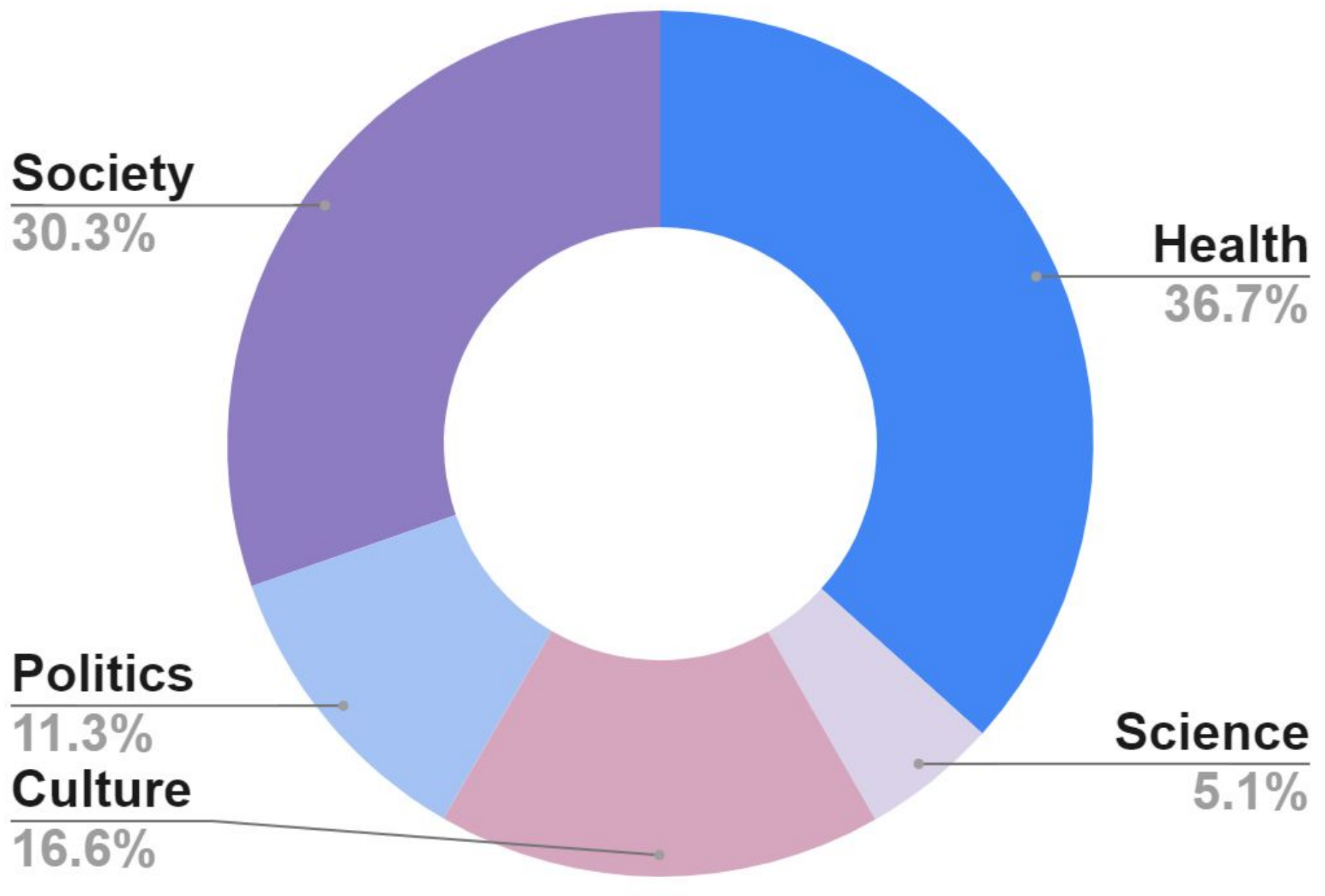}
    % \vspace{-2mm}
    \caption{Distributions of domains. Each instance is categorized into five different domains.}
    \label{fig:ratio}
    \vspace{-5mm}
\end{figure}

\subsection{Claim Labeling}  
The major challenge of constructing a non-English dataset is extracting a claim and its veracity from a fact-checking article usually requires human efforts. Unlike fact-checking articles in English, many non-English articles (e.g.\ Chinese,  Hindi, Filipino) do not explicitly give the claim and assign the veracity.  Therefore, extracting the claim, which can appear in the title, or anywhere in the article, requires manual efforts. Before labeling the claim, we need to extract them from the fact-checking articles.  When performing claim extraction, annotators need to read the fact-checking article first, then identify the claim. They are encouraged to select sentences directly from the article. However, resulted sentences may not be complete, which means they do not provide enough context for fact-checking. One common case is that the sentence describing a fact often lacks the time stamp, or the location. For example, the claim ``Price of pork increases dramatically due to the African swine fever.'' is factual in 2020 but non-factual in 2021.  Therefore, annotators are asked to complete the claim by adding missing content to ensure the claim to be standalone for later verification\footnote{More annotation details are provided in appendix.}. 
Another issue is that Chinese fact-checkers tend to use rhetorical questions to express non-factual claims. To alleviate the bias that the factuality of a claim can be decided by its surface form, annotators are required to paraphrase the questions into declarative statements.

Next, annotators are required to label the extracted claims. English fact-checking articles often provide different truth-rating scales, such as false, mostly false and mixture, while many non-English counterparts do not have such taxonomies. Therefore, annotators need to label the extracted claim based on the understanding of the fact-checking article. Journalism researchers showed that fine-grained labels are often assigned inconsistently due to subjectivity~\citep{uscinski2013epistemology, Lim2018CheckingHF}. Therefore, we chose to follow previous efforts~\citep{Thorne2018FEVERAL, Hanselowski2019ARA} by adopting three types of labels: supported (SUP), refuted (REF) and not enough information (NEI), given the evidence. The distribution of labels in CHEF is shown in Table~\ref{tab:split}. CHEF consists of a majority of refuted claims, as the majority of fact-checking articles aim to debunk non-factual claims.

\begin{figure}
    \centering
    \includegraphics[scale=0.75]{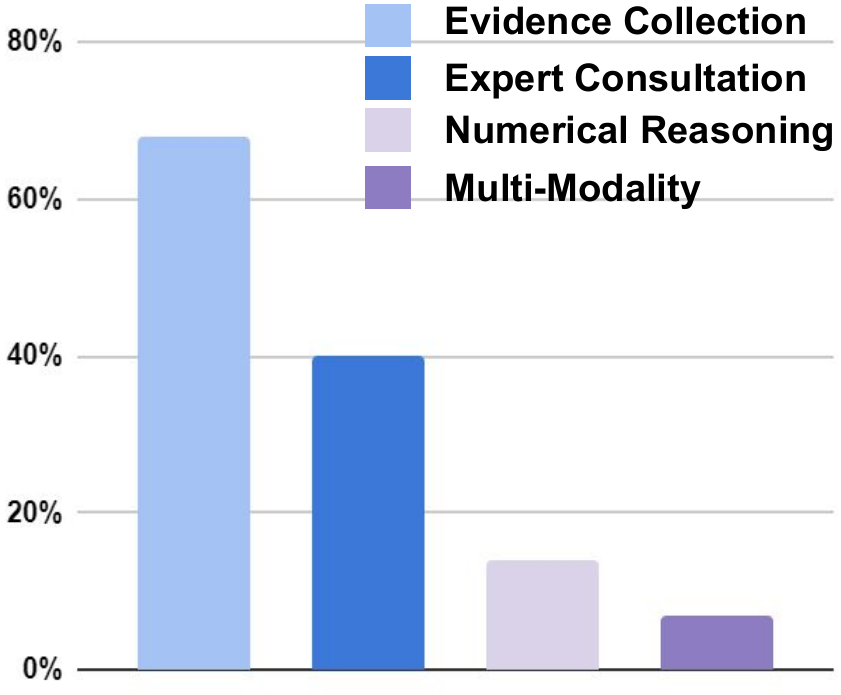}
    \caption{Distributions of challenges. Each instance can have multiple challenges. Evidence collection means finding relevant textual information from web-sources. Expert consultation collects information directly from relevant people. Numerical reasoning requires inference over numbers and multi-modality requires collect and infer over multi-modal evidence.}
    \label{fig:domain}
    \vspace{-5mm}
\end{figure}

\subsection{Evidence Retrieval} 
When verifying a claim, journalists first find information relating to the fact and evaluate it given the collected evidence. As shown in Figure~\ref{fig:ratio}, the biggest challenge of verifying a claim is to collect relevant evidence. In order to validate real-world claims, we chose to manually extract evidence from web-sources. We have two measures to ensure the reliability of the evidence. Firstly, we maintained a list of misinformation and disinformation websites, all search results from these websites will be filtered out. Secondly, we required the annotators to manually select evidence sentences from the search results. In order to collect evidence from the web-sources, we first submitted each claim as a query to the Google Search API by following~\citet{augenstein2019multifc} and~\citet{gupta2021}. The ten most highly ranked search results are retrieved. For each result, we saved the search rank, URL, time stamp and document. Then we filtered out results from fact-checking websites to prevent the answer from being trivially found. Next, annotators were asked to select sentences from the resulted documents. To maintain a balance between keeping relevant and removing irrelevant information, we followed~\citet{Thorne2018FEVERAL} and~\citet{Hanselowski2019ARA} that up to five sentences were selected as evidence. Before deciding which sentences should be selected, annotators were required to answer auxiliary questions, such as ``Whether selected sentences provide sufficient information for factual verification?'' They were encouraged to select the five most relevant sentences, but they were allowed to pick less when relevant sentences are not available. A small fraction (5.6\%) of instances do not have any relevant evidence, and we chose to discard them.

\begin{table}
\centering
\scalebox{0.9}{
\begin{tabular}{lcccc}
\toprule
Split & SUP & REF & NEI & Total \\
\midrule
Train & 2,877 & 4,399 & 776 & 8,002 \\
Dev & 333 & 333 & 333 & 999 \\
Test & 333 & 333 & 333 & 999 \\
\midrule
\midrule
\multicolumn{4}{l}{Avg \#Words in the Claim} & 28 \\
\multicolumn{4}{l}{Avg \#Words in the Google Snippets} & 68 \\
\multicolumn{4}{l}{Avg \#Words in the Evidence Sentences} & 126 \\
\multicolumn{4}{l}{Avg \#Words in the Source Documents} & 3,691 \\
\bottomrule
\end{tabular}}
% \vspace{-2mm}
\caption{Dataset split sizes and statistics for CHEF.}
\label{tab:split}
\vspace{-2mm}
\end{table}

\subsection{Data Validation} 
To ensure the annotation consistency, we conducted an additional 5-way inter-annotator agreement and manual validation. For inter-annotator agreement, we randomly selected 3\% (n = 310) of claims to be annotated by 5 annotators. We calculated the Fleiss $K$ score~\citep{Fleiss1971MeasuringNS} to be 0.74, which is comparable with 0.68 reported in \citet{Thorne2018FEVERAL} and 0.70 in \citet{Hanselowski2019ARA}. In order to verify if evidence sentences provide sufficient information, we chose another 310 instances. The second group of annotators were required to assign the labels based on the evidence sentences. We found that 88.7\% of the instances were labeled correctly and 83.6\% of them provided sufficient information to determine the veracity.  Finally, as shown in Table~\ref{tab:split}, we partitioned the dataset CHEF into training, development and test sets. Our development and test sets have balanced class distributions. Each claim is paired with Google snippets, evidence sentences and source documents.

\section{Baseline Systems}
\label{sec:baseline}

Unlike previous natural datasets, CHEF requires the system to first retrieve the evidence sentences from the documents, then predict the veracity based on the evidence. Therefore, we design two types of baselines: pipeline and joint systems.

\subsection{Pipeline System}
\label{ssec:pipeline}

The pipeline system treats evidence retrieval and veracity prediction as two independent steps.

\subsubsection{Evidence Retrieval}
% \noindent \textbf{Evidence Retrieval:}
Given the claim and documents, this step aims to select the most relevant sentences from documents as evidence, which can be viewed as a ranking problem. Thus, we adopt the following models:

% \noindent \textbf{Surface Ranker}: 
\paragraph{Surface Ranker}
Following retrieval models designed for synthetic datasets~\citep{Thorne2018FEVERAL,jiang2020hover, Aly2021FEVEROUSFE}, 
We use TF-IDF to sort the most similar sentences first and tune a cut-off using validation accuracy on the dev set.

% \noindent \textbf{Semantic Ranker}:
\paragraph{Semantic Ranker}
Inspired by~\citet{Nie2019CombiningFE} and~\citet{Liu2020KernelGA}, we choose semantic matching based on BERT~\citep{Devlin2019BERTPO} pre-trained on Chinese corpus~\citep{Wolf2020TransformersSN}. 
The cosine similarity scores between the embedding of the claim and the embeddings of other sentences in the document are used for ranking.

% \noindent \textbf{Hybrid Ranker}: 
\paragraph{Hybrid Ranker} 
Since semantic encoding is complementary to surface form matching, they can be combined for better ranking. Following~\citet{Shaar20}, we use the rankSVM, based on the feature sets of rankings returned by TF-IDF and the similarity scores computed with BERT.
    
% \noindent \textbf{Google Snippets}:
\paragraph{Google Snippets}
As discussed in Section~\ref{sec:related}, existing natural datasets~\citep{augenstein2019multifc,gupta2021} do not require the system to retrieve the evidence sentences from the documents. Instead, they used summary snippets returned by the Google Search Engine as evidence. We also include this type of evidence for comparisons.

\subsubsection{Veracity Prediction}
% \noindent \textbf{Veracity Prediction:}
After retrieving the evidence sentences, veracity prediction aims to predict the label of the given claim. We implement the following classifiers:

% \noindent \textbf{BERT-Based Model}: 
\paragraph{BERT-Based Model} 
Following~\citet{jiang2020hover} and \citet{vitaminc2021}, we use a multi-layer perceptron with embeddings from BERT as the classifier. The embeddings of claim and retrieved evidence are concatenated as the input. The model performs the classification based on the output representation of the \texttt{CLS} token.

% \noindent \textbf{Attention-Based Model}: 
\paragraph{Attention-Based Model}
Following~\citet{gupta2021}, we first extract the output embedding of the \texttt{CLS} token of each selected evidence and calculate relevance weights with the claim through dot product attention. Then we feed the concatenated claim and weighted evidence into the BERT-based classifier.

% \noindent \textbf{Graph-Based Model}: 
\paragraph{Graph-Based Model}
Recent efforts~\citep{zhou2019gear,Zhong2020ReasoningOS,Liu2020KernelGA} showed that graphs help to capture richer interactions among multiple evidence for fact-checking. We adopt the Kernel Graph Attention Network~\citep{Liu2020KernelGA} for veracity prediction. The evidence graph is constructed based on the claim and evidence sentences, then node and edge kernels are used to conduct fine-grained evidence propagation. The updated node representations are used to calculate the claim label probability.

\subsection{Joint System}
\label{ssec:joint}

Evidence retrieval in the pipeline system could not solicit optimization feedback obtained from veracity prediction. In order to optimize two steps jointly, we proposed to model the evidence retrieval as a latent variable. The joint system contains two modules: a latent retriever and a classifier. For the classifier, we used the same models described in Section~\ref{ssec:pipeline}. Latent retriever labels each sentence in the documents with a binary mask. Sentences labeled with 1 are selected as the evidence, while sentences labeled with 0 will be neglected. 

\subsubsection{Latent Retriever}

We built the latent retriever based on the Hard Kumaraswamy distribution~\citep{bastings2019interpretable}, which gives support to binary outcomes and allows for reparameterized gradient estimates\footnote{Please refer to the detailed derivations in \citet{bastings2019interpretable}.}. We first stretch the Kumaraswamy distribution~\citep{kumaraswamy1980generalized} to include $0$ and $1$ by the support of open interval ${(l,r)}$ where ${l<0}$ and ${r>1}$, defined as $K^{'} \sim \operatorname{Kuma}(a, b, l, r)$ with CDF:
\begin{align}
F_{K^{'}}(k^{'} ; a, b, l, r)=F_{K^{'}}((k^{'}-l) /(r-l) ; a, b)
\end{align}
A sigmoid function $k^{''}$=$\min(1, \max (0, k^{'}))$ is used to rectify random variables into the closed interval ${[0,1]}$, denoted by $K^{''}\sim\operatorname{HardKuma}(a, b, l, r)$ and $k^{''}$=$s(u;a, b, l, r)$ for short. Note that we map all negative values ${k^{'} \in (l,0]}$ into $k^{''}$=$0$ and $k^{'}$ $\in [1,r)$ into $k^{''}$=$1$ deterministically, so the sets whose masses under $\operatorname{Kuma}(k^{'}|a, b, l, r)$ are available in the closed form:
\begin{equation}
\begin{aligned} 
\mathbb{P}(K^{''}=0)&=F_{K}(\frac{-l}{r-l} ; a, b)\\
\mathbb{P}(K^{''}=1)&=1-F_{K}(\frac{1-l}{r-l} ; a, b)
\end{aligned} 
\end{equation}
Given  source documents $D$, the latent retriever selects relevant sentences as evidence that can be used to predict the veracity for the claim $c$. For the $i$-th sentence $x_{i} \in D$, we obtain the sentence-level embedding $\mathbf{h}_{i}$ based on a BERT encoder by using the \texttt{CLS} token. Then we can calculate the latent selector $k^{''}_{i}$ by:
\begin{equation}
\begin{aligned} 
k^{''}_{i} &= s(u_{i};a_{i}, b_{i}, l, r),\\
a_{i}=f_{a}\left(\mathbf{h}_{i} ; \phi_{a}\right) \quad b_{i}&=f_{b}\left(\mathbf{h}_{i} ; \phi_{b}\right)  \quad
u_{i}  \sim \mathcal{U}(0,1)
\end{aligned}
\end{equation}
where ${f_{a}\left(\cdot; \phi_{a}\right)}$ and ${f_{b}\left(\cdot; \phi_{b}\right)}$ are feed-forward networks with $\operatorname{softplus}$ outputs ${a_{i}}$ and ${b_{i}}$. ${s(\cdot)}$ turns the uniform sample ${u_{i}}$ into the latent selector $k_{i}^{''}$. Next, we use the sampled $k^{''}_{i}$ to modulate inputs to the classifier for veracity prediction:
\begin{equation}
f_{F}\left(k_{i}^{''}\cdot\mathbf{h}_{i},\mathbf{c};\theta_{F}\right)
\end{equation}
where $\mathbf{c}$=$f_{\theta^{'}}(c)$ denotes the embedding for the given claim $c$ obtained by using the \texttt{CLS} token through a BERT encoder. ${f_{F}\left(\cdot; \theta_{F}\right)}$ represents the classifier (e.g.\ graph-based model). The joint system can be optimized by gradient estimates of ${\mathcal{E}(\phi, \theta)}$ via Monte Carlo sampling from:
\begin{equation}
\mathcal{E}(\phi, \theta)=\mathbb{E}_{\mathcal{U}(0, 1)}\left[\log P\left(y \mid X, s_{\phi}(u, X), \theta\right)\right]
\end{equation}
where $y$ is the label of veracity and $k^{''}_{i}=s_{\phi}(u, X)$ abbreviate the transformation from uniform samples to $\operatorname{HardKuma}$ samples. 

\subsubsection{More Baselines}
Apart from the proposed system, we include following baselines for comprehensive comparisons:

% \noindent \textbf{Reinforce}: 
\paragraph{Reinforce}
Instead of using gradient-based training, we follow~\citet{lei2016rationalizing} by assigning a binary Bernoulli variable to each evidence sentence. Because gradients do not flow through discrete samples, the evidence retriever is optimized using REINFORCE~\citep{williams1992simple}. A $L_{0}$ regularizer is used to impose sparsity-inducing penalties.

% \noindent \textbf{Multi-task}:
\paragraph{Multi-task}
We also adopt the multi-task learning method proposed by~\citet{Yin2018}, which is the state-of-the-art joint model on FEVER dataset~\citep{Thorne2018FEVERAL}. The model predicts a binary vector that indicates the subset of sentences as evidence, and a one-hot vector indicates the veracity of the claim. The overall training loss is the sum of these two prediction losses.

\begin{table*}
\centering
\scalebox{0.84}{
\begin{tabular}{ccccccccc}
\toprule
\multicolumn{3}{c}{\multirow{2}{*}{System / Evidence}} & \multicolumn{2}{c}{BERT-Based Model$^{1}$} & \multicolumn{2}{c}{Attention-Based Model$^{2}$} & \multicolumn{2}{c}{Graph-Based Model$^{3}$} \\ \cmidrule(lr){4-5} \cmidrule(lr){6-7} \cmidrule(lr){8-9} 
& & & Micro F1 & Macro F1 & Micro F1 & Macro F1 & Micro F1 & Macro F1  \\
\midrule
\multirow{6}{*}{Pipeline} & \multicolumn{2}{|l|}{No Evidence} & 54.46\small±2.89 & 52.49\small±2.44 & 54.87\small±1.95 & 53.47\small±2.82 & --- & --- \\
& \multicolumn{2}{|l|}{Snippets} &\cellcolor{blue!15}62.07\small±2.55 &\cellcolor{blue!15}60.61\small±2.96  & \cellcolor{blue!15}62.42\small±2.31 & \cellcolor{blue!15}60.24\small±2.56  & \cellcolor{blue!15}62.78\small±1.70 & \cellcolor{blue!15}61.06\small±2.59  \\
& \multicolumn{2}{|l|}{Surface Ranker}  & \cellcolor{green!15}63.17\small±1.67 & \cellcolor{green!15}61.47\small±2.02  & \cellcolor{green!15}63.77\small±1.89  & \cellcolor{green!15}62.65\small±2.32  & \cellcolor{green!15}64.58\small±1.45 & \cellcolor{green!15}61.46\small±1.72\\
& \multicolumn{2}{|l|}{Semantic Ranker} & \cellcolor{green!15}\textbf{63.47\small±1.71} & \cellcolor{green!15}\textbf{61.94\small±1.66} & \cellcolor{green!15}\textbf{63.95\small±1.46} & \cellcolor{green!15}\textbf{62.80\small±1.33}  &\cellcolor{green!15}\textbf{64.67\small±1.54} & \cellcolor{green!15}62.28\small±1.50 \\
& \multicolumn{2}{|l|}{Hybrid Ranker} & \cellcolor{green!15}63.29\small±1.65  & \cellcolor{green!15}61.80\small±2.31  & \cellcolor{green!15}63.48\small±1.22 & \cellcolor{green!15}62.74\small±1.30 & \cellcolor{green!15}64.37\small±1.66 & \cellcolor{green!15}\textbf{62.58\small±1.43}\\ 
\midrule
\multicolumn{1}{c}{\multirow{6}{*}{Joint}} & \multicolumn{1}{|l}{\multirow{2}{*}{Reinforce$^{4}$}} &
\multicolumn{1}{|l|}{Snippets} &\cellcolor{blue!15} 63.76\small±1.52 & \cellcolor{blue!15}61.74\small±1.88 &\cellcolor{blue!15}64.06\small±1.76 &\cellcolor{blue!15}61.97\small±1.04   &\cellcolor{blue!15}65.77\small±1.23 &\cellcolor{blue!15}62.34\small±1.11 \\   
&\multicolumn{1}{|l}{} & \multicolumn{1}{|l|}{Documents} &\cellcolor{green!15}64.37\small±1.65 & \cellcolor{green!15}62.46\small±1.72 & \cellcolor{green!15}64.86\small±1.83 & \cellcolor{green!15}62.66\small±1.32   & \cellcolor{green!15}66.58\small±1.45 & \cellcolor{green!15}63.47\small±1.58 \\
\cmidrule(lr){2-3} 
 & \multicolumn{1}{|l}{\multirow{2}{*}{Multi-task$^{5}$}} &
\multicolumn{1}{|l|}{Snippets} &\cellcolor{blue!15} 62.78\small±1.41 & \cellcolor{blue!15}61.98\small±2.59 &\cellcolor{blue!15}64.43\small±1.72 &\cellcolor{blue!15}61.58\small±1.34   &\cellcolor{blue!15}66.21\small±1.57 &\cellcolor{blue!15}63.15\small±1.46 \\   
& \multicolumn{1}{|l}{} & \multicolumn{1}{|l|}{Documents} &\cellcolor{green!15} 65.02\small±1.46 & \cellcolor{green!15}63.12\small±1.78 & \cellcolor{green!15}65.45\small±1.59 & \cellcolor{green!15}62.94\small±2.03   & \cellcolor{green!15}67.46\small±1.72 & \cellcolor{green!15}64.31\small±1.81 \\
\cmidrule(lr){2-3} 
 &\multicolumn{1}{|l}{\multirow{2}{*}{Latent}} &
\multicolumn{1}{|l|}{Snippets} &\cellcolor{blue!15} 64.45\small±1.68 & \cellcolor{blue!15}62.52\small±2.23 &\cellcolor{blue!15}65.73\small±1.75 &\cellcolor{blue!15}63.44\small±1.68   &\cellcolor{blue!15}67.81\small±1.74 &\cellcolor{blue!15}64.34\small±1.57 \\   
& \multicolumn{1}{|l}{} & \multicolumn{1}{|l|}{Documents} &\cellcolor{green!15} \textbf{66.77\small±1.43} & \cellcolor{green!15}\textbf{64.65\small±1.74} & \cellcolor{green!15}\textbf{67.62\small±1.48} & \cellcolor{green!15}\textbf{64.81\small±1.26}   & \cellcolor{green!15}\textbf{69.12\small±1.13} & \cellcolor{green!15}\textbf{65.26\small±1.67} \\
\midrule
\midrule
\multirow{1}{*}{Pipeline} &
\multicolumn{2}{|l|}{Gold Evidence} & \textbf{78.99\small±0.82} & \textbf{77.62\small±1.02}  & \textbf{79.18\small±1.07} & \textbf{78.36\small±1.40} & \textbf{79.84\small±1.24} & \textbf{78.47\small±1.17}  \\

\bottomrule
\multicolumn{9}{c}{\citet{vitaminc2021}$^{1}$, \citet{gupta2021}$^{2}$, \citet{Liu2020KernelGA}$^{3}$, \citet{lei2016rationalizing}$^{4}$, \citet{Yin2018}$^{5}$}
\end{tabular}}
\vspace{-2mm}
\caption{Results of pipeline and joint systems on CHEF.}
\label{tab:verification}
\vspace{-2mm}
\end{table*}

\section{Experiments and Analyses}
\label{sec:experiments}

\subsection{Experimental Setup}
Following~\citet{augenstein2019multifc}, we computed the Micro F1 and Macro F1 as the evaluation metric. We further reported the mean F1 score and standard deviation by using 5 models from independent runs. For the pipeline system, we used 6 different evidence settings, including evidence sentences retrieved by surface ranker, semantic ranker and hybrid ranker, Google snippets, gold evidence and without using any evidence. For the joint system, we used 2 types of evidence: Google snippets and source documents, where the latent retriever can select sentences from. We used three classifiers for both systems, BERT-based~\citep{vitaminc2021}, attention-based~\citep{gupta2021} and graph-based models~\citep{Liu2020KernelGA}.

The hyper-parameters are chosen based on the development set. In the evidence retrieval step of the pipeline system, we set the retrieved evidence obtained from TF-IDF to be more than $5$ words for surface ranker. We use the BERT default tokenizer with max-length as $256$ to preprocess data for semantic ranker. We use the default parameters in sklearn.svm.LinearSVC with RBF kernel for hybrid ranker.

In the veracity predication step of the pipeline system, we use the BERT default tokenizer with max-length as $256$ and pretrained BERT-base-Chinese as the initial parameter to encode claim and evidence\footnote{\url{https://huggingface.co/}}. For BERT-based model, the fully connected network for classification is defined with layer dimensions of ${{h_{R}}}$-${h_{R}/2}$-verification\_labels, where ${{h_{R}}=768}$. We use BertAdam~\citep{Devlin2019BERTPO} with $5e{-6}$ learning rate, warmup with $0.1$ to optimize the cross entropy loss and set the batch size as $16$. For attention-based model, we use BertAdam with $2e{-5}$ learning rate, warmup with $0.1$ to optimize the cross entropy loss and set the batch size as $8$. For graph-based model, we use BertAdam with $5e{-5}$ learning rate, warmup with $0.1$, batch size with $16$, dropout with $0.6$ and kernel size with $21$.

For the joint system, we use Adam~\citep{kingma2014adam} with $5e{-5}$ learning rate, learning rate decay with $0.5$ to optimize the cross entropy loss. We set the batch size as $32$. The fully-connected networks ${f_{a}\left(\cdot; \phi_{a}\right)}$ and ${f_{b}\left(\cdot; \phi_{b}\right)}$ for two parameters ${a_{i}}$ and ${b_{i}}$ are defined with layer dimensions of ${{h_{R}}=768}$. We set the dropout rate as $0.5$.

\subsection{Main Results}

\paragraph{Pipeline System:} 
According to Table~\ref{tab:verification}, pipeline systems with evidence including Google snippets, sentences returned by rankers and gold evidence consistently outperform systems without using evidence. These results confirm that evidence plays an important role in verifying real-world claims. On the other hand, systems with retrieved sentences achieve higher scores than systems with Google snippets. Specifically, systems with gold evidence significantly outperform the ones with Google snippets, indicating information that is necessary for verification is missing in the snippets. Moreover, systems with retrieved evidence are more robust in terms of standard deviation. We hypothesize the reason is that irrelevant information is presented in the snippets. When comparing with different rankers, we observed that using contextualized representations to measure the similarity (Semantic Ranker) is generally better than exact string match (Surface Ranker). However, there still exists a large performance gap between the pipeline system with semantic ranker and the system with gold evidence. One potential solution is to develop better retrieval models based on the supervision signal of gold evidence provided by CHEF. Given the evidence sentences, graph-based models tend to have higher scores than BERT-based and attention-based models, which shows the effectiveness of leveraging graph structure to synthesize multiple evidence.

% \noindent \textbf{Joint System:}
\paragraph{Joint System:} 
Similar to the pipeline systems, joint systems that retrieve evidence sentences from documents achieve better F1 scores than directly use the summary snippets. In order to verify real-world claims, it is necessary to train fact-checking systems that learn how to effectively retrieve evidence sentences from full documents on web pages. In addition, joint system outperforms pipeline system consistently with both Google snippets and source documents as inputs. For example, latent retriever with Google snippets are able to achieve an average 2.74\% and 1.77\% Micro/Macro F1 boost compared with the pipeline systems with the same type of evidence. 
We attribute the consistent improvement of joint system to the explicit feedback to the evidence retrieval via gradient estimation on veracity prediction. Another advantage of the joint system is that the latent evidence retriever is able to dynamically select relevant sentences from documents for each instance, while rankers return a fixed number of evidence.

Compared with the reinforce and multi-task methods, the proposed latent retriever achieves 1.41\% and 1.98\% higher F1 on average with Google snippets and source documents as inputs across various classifiers. When considering standard deviation, reinforce is less robust. We believe the main reason is that latent retriever facilitate training through differentiable binary variables, which leads to robust and generalized model that exhibits small variance over multiple runs.

\subsection{Analysis and Discussion}
In this section, we further provide fine-grained analyses for baseline systems on CHEF. For brevity, we abbreviate pipeline systems with Google Snippets, Surface Ranker, Semantic Ranker, Hybrid Ranker as \textbf{GS}, \textbf{Sur}, \textbf{Sem}, \textbf{Hyb}, while joint systems with Google snippets and source documents as inputs as \textbf{JG} and \textbf{JS}, respectively. All results are reported based on the BERT-based model. We further provide case study and error analysis on CHEF. Due to limited spaces, we attach them in the appendix.

\paragraph{Effect of Evidence:} 
In Table~\ref{tab:evidence}, we varied the numbers of evidence retrieved and reported the Macro F1 on the test set. The fluctuation results indicate that both quantity and quality of retrieved evidence affect the performance. Using fewer evidence will lead to incomplete coverage, which may not provide sufficient information to verify the claims. On the other hand, incorporating more evidence may introduce irrelevant sentences thus propagate errors to veracity prediction. In general, systems with 5 evidence sentences achieves the best performance except the joint system with source documents as inputs. We believe the reason is that the latent retriever maintains a better balance between keeping relevant and removing irrelevant sentences, which helps to achieve higher scores with more evidence sentences.

\begin{table}
\centering
\scalebox{0.9}{
\begin{tabular}{lcccc|cc}
\toprule
\textbf{\#E} & \textbf{GS} & \textbf{Sur} & \textbf{Sem} & \textbf{Hyb} & \textbf{JG} & \textbf{JS} \\
\midrule
1 & 55.24 & 55.67 & 56.04 & 56.72 & 56.98 & 57.54 \\
3 & 58.69 & 59.24 & 59.52 & 59.18 & 59.89 & 61.45 \\
5 & \textbf{60.61} & \textbf{61.47} & \textbf{61.94} & \textbf{61.80} & \textbf{62.12} & 64.65 \\
10 & 59.12 & 60.20 & 60.37 & 61.24 & 61.86 & \textbf{64.73} \\
15 & 55.72 & 56.31 & 56.56 & 57.08  & 58.69 & 59.11\\
\bottomrule
\end{tabular}}
\vspace{-2mm}
\caption{Effects of evidence: Macro F1 scores on the test set are reported. \#E indicates the number of evidence.}
\label{tab:evidence}
\vspace{-2mm}
\end{table}

\begin{figure}
    \centering
    \includegraphics[scale=0.48]{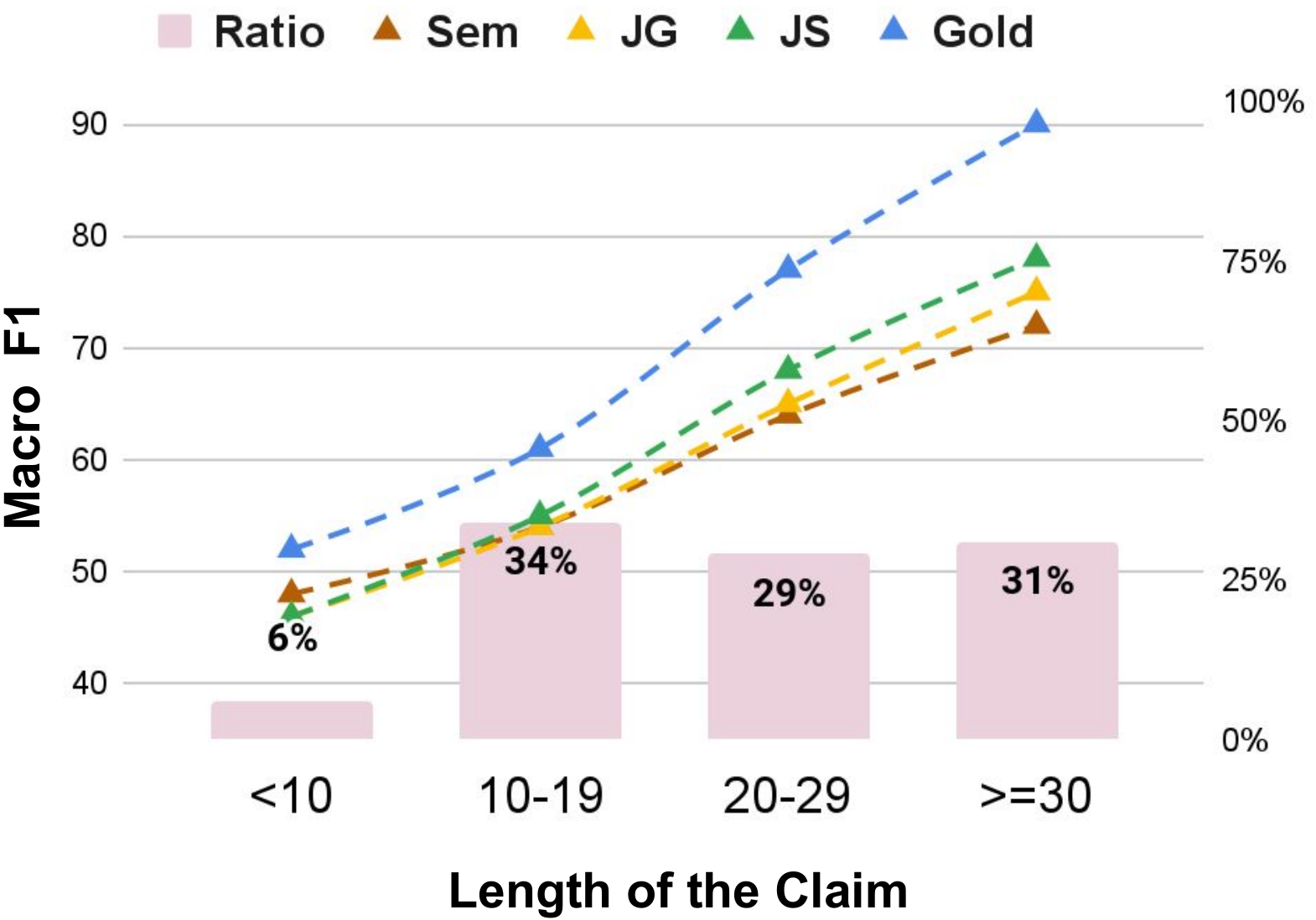}
    \vspace{-3mm}
    \caption{Comparisons against claim lengths.}
    \label{fig:length}
    \vspace{-1mm}
\end{figure}

\paragraph{Performance against Claim Length:}
% \noindent\textbf{Performance against Claim Length:} 
We partitioned the test set into 4 classes (\textless 10, 10-19, 20-29, ${\geq}$30) based on lengths of the claims and reported the Macro F1 score. For clarity, we choose the best reported pipeline system with semantic ranker to compare with joint systems. As shown in Figure~\ref{fig:length}, most claims are longer than 10 words. Performance of the systems on short claims (e.g.\ \textless 10) is lower than other. One reason is that such claims do not contain sufficient information to retrieve evidence and to be verified, based on the observation that the performance of all the systems improve as the length of the claim increase. In general, the joint system outperforms the pipeline system against various claim lengths.

\begin{figure}
    \centering
    \includegraphics[scale=0.54]{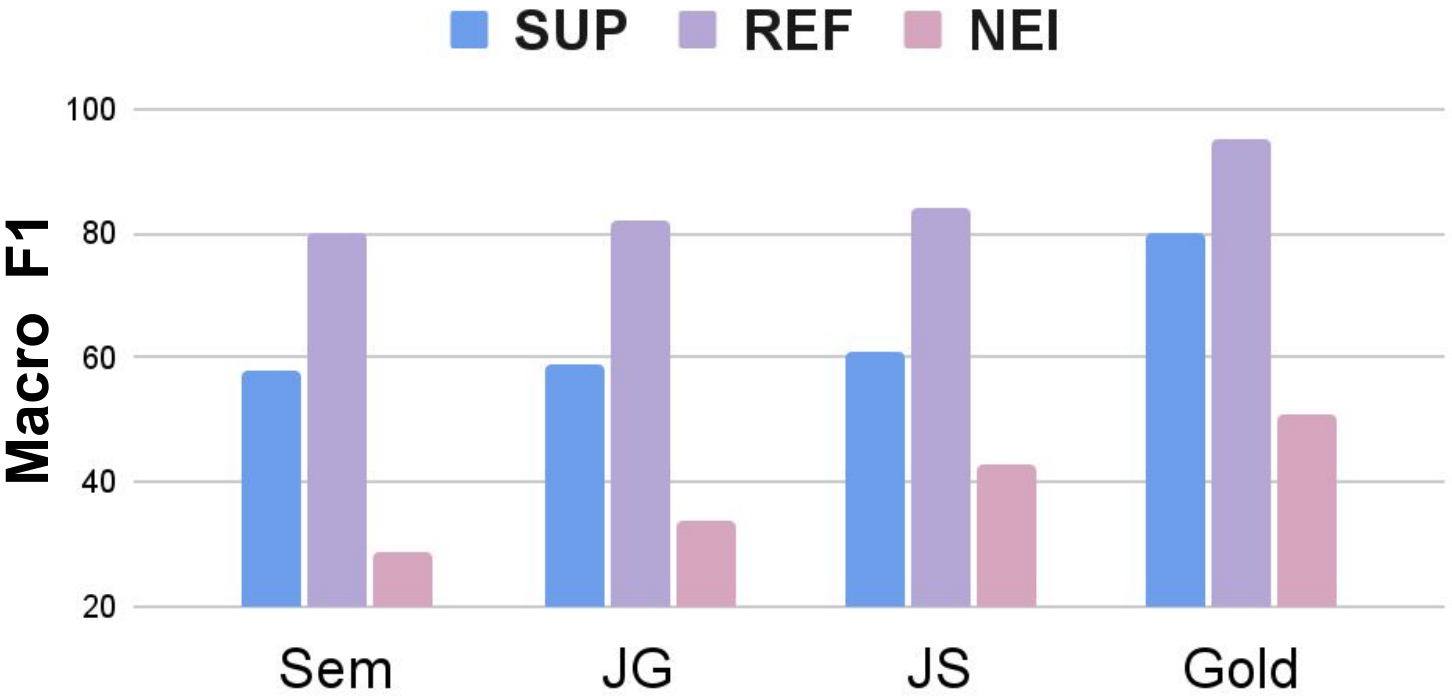}
    \vspace{-2mm}
    \caption{Per-class results.}
    \label{fig:class}
    \vspace{-1mm}
\end{figure}
\begin{figure}
    \centering
    \includegraphics[scale=0.44]{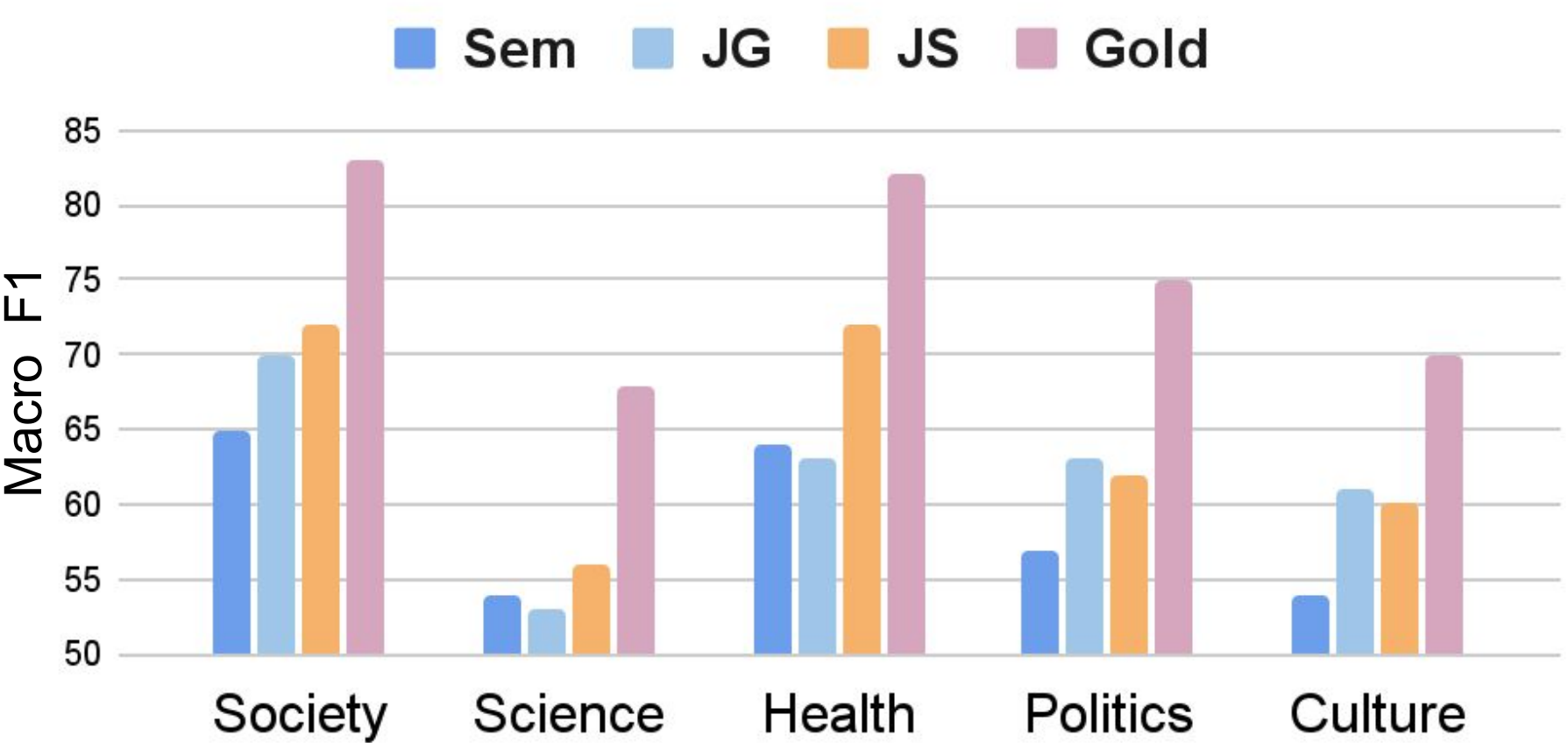}
    \vspace{-3mm}
    \caption{Per-domain results.}
    \label{fig:domain}
    \vspace{-1mm}
\end{figure}

\paragraph{Performance against Classes and Domains:}
% \noindent\textbf{Performance against Classes and Domains:} 
As CHEF is constructed based on real-world claims, most of them are non-factual claims verified by fact-checking websites. Such an imbalance issue poses a challenge to the fact-checking system. Figure~\ref{fig:class} shows the performance of models for different veracity labels. The scores of minor classes are much lower than the majority class. This reflects the difficulty of judging SUP and NEI. Informative evidence helps to alleviate this issue. For example, the pipeline system with gold evidence achieves significant improvement on predicting NEI labels when comparing with the system with semantic ranker. Figure~\ref{fig:domain} shows the performance of different domains. Claims from science, politics and culture domains have fewer training instances as most claims in the dataset focus on the society and public health topics. Again, retrieving informative evidence sentences (JS and Gold) from full documents is beneficial to this data sparsity issue.

\section{Conclusion}
\label{sec:conclusion}

We constructed the first Chinese dataset for evidence-based fact-checking. Further, we have discussed the annotation methods and shared some of the insights obtained that will be useful to other non-English annotation efforts. To evaluate the challenge CHEF presents, we have developed established baselines and conducted extensive experiments. We show that the task is challenging yet feasible. We believe that CHEF will provide a stimulating challenge for automatic fact-checking.

\section{Acknowledgement}
We thank the reviewers for their valuable comments. The work was supported by the National Key Research and Development Program of China (No. 2019YFB1704003), the National Nature Science Foundation of China (No. 62021002 and No. 71690231), NSF under grants III-1763325, III-1909323, III-2106758, SaTC-1930941, Tsinghua BNRist and Beijing Key Laboratory of Industrial Bigdata System and Application.
\section{Ethical Consideration}
Datasets have been collected in a manner that is consistent with the terms of use of any sources and the intellectual property. For each annotator, we compensate based on the number and quality of annotated sentences. More details of our datasets are depicted in Section~\ref{sec:construction}.

\bibliography{custom}
\bibliographystyle{acl_natbib}

\newpage
\appendix
\newpage

\section{Supplement Materials}

\subsection{CHEF annotation guidelines}
\begin{CJK*}{UTF8}{gbsn}
\subsubsection{\textbf{声明抽取和规范化的指引} Guidelines for claim extraction and normalization:}
标注者首先需要认真阅读事实验证的文章，然后使用一到两句话来概括这篇文章描述的事件作为声明。请标注者直接使用文章中的句子作为声明，比如文章的标题，或者首段的前几句话都可能是这篇文章需要验证的事实。如果没法抽取出相应的句子，可以使用自己的语言总结文章来撰写声明。在撰写声明的时候，有以下注意事项：
\begin{itemize}
\item每个声明必须完整。
\item每个声明不应该存在事实验证偏差。
\item每个声明不应该存在信息泄露。
\end{itemize}
请仔细阅读以下详细指引和相应的规范化例子:
\begin{itemize}
\item声明中描述的事件缺乏必要的细节,比如：时间和地点。标注者需要加上这些细节让声明完整，才能被验证。比如：今年共有12.08万人参加中考，但招生计划只有4.3万。需要改写声明为：2019年，共有12.08万人参加成都中考，但招生计划只有4.3万。 
\item声明中存在特殊符号， 需要去除特设符号避免声明中存在偏差。比如：“纯天然”喷雾一喷“秒睡”。需要去除句子中的“”，因为这些特殊符号隐晦地表达了这个声明其实是不实的。模型可以通过特殊符号直接判断一个声明的真实性。这个句子需要改写为：纯天然喷雾一喷秒睡。，这可以避免由于声明中的特殊符号“”带来事实验证偏差。
\item声明中存在信息泄露，需要去除直接指出声明真实性的相关词语。比如：谣言！纯天然喷雾一喷秒睡。句子中使用的“谣言！”已经直接指出该声明是不实的，造成了信息泄露。需要改写声明为：纯天然喷雾一喷秒睡。不能在声明中出现诸如：“谣言”、“错误”，“骗局”等信息泄露词。
\item声明中的反问句需要被改写为陈述句，由于采用反问句形式的声明大部分都是不实的，反问句的形式会造成数据集偏差。比如：别人打了新冠疫苗，我们就可以不打新冠疫苗吗？需要被改写为：别人打了新冠疫苗，我们就可以不打新冠疫苗。
\item不陈述事实的声明需要被丢弃。有两大类的声明是无法进行事实验证的。第一大类为表示推测的声明，比如：明年深圳房价会上涨。第二大类为表示个人意见的声明，比如：我认为特朗普应该连任。
\item声明中如果包含多个声明，需要拆分为多个声明逐一验证。比如：关于新冠疫苗接种的两个事实：第一，别人打了新冠疫苗，自己就可以不打新冠疫苗。其次，新冠疫苗只需要打一针就能具备新冠病毒防护能力。这个声明包括了两个子声明，需要被拆分为：别人打了新冠疫苗，我就可以不打新冠疫苗。第二个声明为：新冠疫苗只需要打一针就能具备新冠病毒防护能力。

\end{itemize}
The annotator first needs to read the fact-checking article carefully, and then use one or two sentences to summarize the event described in this article as a claim. The annotator is encouraged to directly extract the sentences in the article as the claim, such as the title of the article, or the first few sentences in the first paragraph. If the annotator cannot find the sentence that can serve as a claim, you can use your own language to write the claim. When extracting the claim, there are the following considerations:
\begin{itemize}
\item Each claim must be complete.
\item For each claim, explicit bias should be removed.
\item Each claim should not have information leakage.
\end{itemize}
Please read the following detailed guidelines and corresponding normalized examples carefully:
\begin{itemize}
\item If the event described in the claim lacks important details, such as time and location, annotator need to add these necessary metadata to make the claim complete before it can be verified. For example, a total of 120,800 people took the entrance examination this year, but the enrollment plan is only 43,000. The claim needs to be rewritten as follows: In 2019, a total of 120,800 people participated in the Chengdu high school entrance examination, but the enrollment plan was only 43,000.
\item If there exist special symbols in the claim, such symbols that may lead to bias for claim verification should be removed. For example: "Natural spray" helps you "sleep instantly". Quotation marks should be removed in the sentence, as these special symbols implicitly indicates such a claim is non-factual. The model can predict the veracity simply based on the special symbols in the claim. This claim needs to be rewritten as: Natural spray helps you sleep instantly. 
\item Claims contains words that will lead to information leakage should be removed. For example: Rumors! Natural spray helps you sleep instantly. The word "Rumor!" in the claim directly pointed out that the claim is nonfactual, causing information leakage. The word "Rumor!" should be removed. Do not include information leaking words such as "rumors", "errors", "scams", etc. in the claim.
\item Claims used rhetorical questions need to be rewritten into declarative sentences. Since most of the claims in the form of rhetorical question are nonfactual, the form of the rhetorical question exhibits a bias in the dataset. For example: if someone else gets the COVID-19 vaccine, can we not get the vaccine? It needs to be rewritten as: if someone else gets a COVID-19 vaccine, we do not need to get the vaccine..
\item Claims that do not related to factuality should be discarded. There are two major types of claims that cannot be verified. The first category is speculative claims, such as: Shenzhen housing prices will rise next year. The second category is claims expressing personal opinions, such as: I think Donald Trump should be the president.
\item A claim contains multiple statements should be split into multiple claims to be verified one by one. For example, a claim stated that: First, if someone else gets the COVID-19 vaccine, you do not need to get one. Also, the COVID-19 vaccine only needs one shot to protect against the virus. This claim includes two sub-claims. It needs to be split into two claims.
\end{itemize}

\subsubsection{\textbf{声明标注的指引}\quad Guidelines for claim labeling}
标注者在抽取出和规范化声明之后，需要根据事实验证的文章给出的结论，给每个声明打上标签。我们提供了以下三种标签，请选择其中的一种。注意的是，对于大部分为真，部分为真，大部分为为假，部分为假和半真半假的情况，我们统一归类为信息不足：
\begin{itemize}
\item 支持，有充分证据表明这个声明是被证据所支持的。
\item 反对，有充分证据表明这个声明是被证据所反对的。
\item 信息不足，没有足够的证据表明这个声明是被支持还是反对。
\end{itemize}
After extracting and normalizing the claim, annotators needs to label each claim based on the conclusions of the fact-checking article. We provide the following three labels, please choose one of them. Note that for conclusions such as mostly true, partially true, mostly false, partial false and mixture, we consider them as not enough information:
\begin{itemize}
\item Supported, there is sufficient evidence to show that this claim is supported by the evidence.
\item Refuted, there is sufficient evidence to show that this claim is refuted by the evidence.
\item Not enough information, there is not enough evidence to show whether this claim is supported or refuted.
\end{itemize}

\subsubsection{\textbf{证据标注的指引}\quad Guidelines for evidence labelling}
标注者需要阅读规范化过后的声明，事实验证的文章还有搜集到的源文档。标注者首先需要理解文章的验证思路，再从源文档当中直接选择能够作为证据的句子。针对每个声明，标注者最少选择1个，最多选择5个相关的句子作为证据。在选择句子作为证据的时候有以下注意事项：
\begin{itemize}
\item 请标注者选择完整的句子，以句号为结束标志。
\item 选择句子作为证据的条件是，在仅仅基于当前选中的句子作为证据的前提下，能够验证给定的声明。也就是说，选中的句子必须要提供给足够的信息来帮助判断声明的事实性。
\item 如果出现多于5个句子能够作为证据的情况，选择你认为最相关的5个句子；或者能够形成推理逻辑链的句子；或者和事实验证文章推理过程最相似的句子。
\item 如果出现源文档互相矛盾的情况，优先选择支持事实验证文章结论的文档，从中选择相关的句子作为证据。
\item 如果提供的源文档并没有提供足够的证据来验证声明，请报告这条声明。
\item 如果提供的源文档只包含和事实验证文章结论矛盾的证据，请报告这条声明。
\end{itemize}
The annotator needs to read the normalized claim, the fact-checking article and the collected documents. Annotators first need to understand the verification process in the fact-checking article, and then directly select sentences from the sources documents. These selected sentence are used as evidence to verify the claim. For each claim, annotators should select at least 1 and at most 5 relevant sentences as evidence. There are the following considerations when choosing sentences as evidence:
\begin{itemize}
\item Please select a complete sentence which ends with a period.
\item When selecting sentences as evidence, the annotator should consider given the selected sentences if the given claim can be verified. In other words, the selected sentences must provide sufficient information to predict the factuality of the given claim.
\item If there are more than 5 sentences that can be used as evidence, choose the 5 sentences that you think are the most relevant; or the sentences that can form a reasoning chain for verification; or the sentence that is most similar to the reasoning process of the fact-checking article.
\item If there are conflicting source documents, the documents that support the conclusion of the fact-checking article should be considered, and the most relevant sentences in these documents are selected as evidence.
\item If the source documents do not provide sufficient evidence to verify the statement, please report the claim.
\item If the source documents only contains evidence that contradicts the conclusion of the fact-checking article, please report this claim.
\end{itemize}

\subsubsection{\textbf{数据验证的指引}\quad Guidelines for data validation}
给定一个声明和搜集到的证据句子，标注者需要根据证据去判断这个声明的真实性。如果标注者认为提供的声明缺失重要信息，或者是不可读的，请报告该条声明。我们提供了以下三种标签，请选择其中的一种：
\begin{itemize}
\item 支持，有充分证据表明这个声明是被证据所支持的。
\item 反对，有充分证据表明这个声明是被证据所反对的。
\item 信息不足，没有足够的证据表明这个声明是被支持还是反对。
\end{itemize}
Given a claim and the evidence sentences, the annotator needs to label the factuality of the claim based on the evidence. If the annotator believes that the given claim lacks important information or is unreadable, please report the claim. We provide three kinds of labels, please choose one of them:
\begin{itemize}
\item Supported, there is sufficient evidence to show that this claim is supported by the evidence.
\item Refuted, there is sufficient evidence to show that this claim is refuted by the evidence.
\item Not enough information, there is not enough evidence to show whether this claim is supported or refuted.
\end{itemize}

\subsubsection{\textbf{判断声明领域的指引}\quad Guidelines for determining claim domain}
标注者需要阅读声明，根据给出的五个领域判断声明属于哪个领域：
\begin{itemize}
\item政治：主要是关于国际与国内政治等方面的声明。
\item公卫：主要是关于公共卫生方面的声明，比如有关新冠病毒，人体健康，食品安全等方面。
\item科学：主要是关于自然科学和工程技术等方面的声明。
\item文化：主要是关于历史，人文，娱乐，体育等方面的声明。
\item社会：主要是除了上述四类，社会生活方面的声明。
\end{itemize}
The annotator needs to read the claim and determine which domain the claim belongs to based on the five domains given:
\begin{itemize}
\item Politics: Claims mainly focus on international and domestic politics.
\item Health: Claims mainly focus on public health, including topic related to COVID-19, health care, food safety, etc.
\item Science: Claims mainly focus on natural science and technology.
\item Culture: Claims mainly focus on history, humanities, entertainment, sports, etc.
\item Society: Claims not related on the above four categories, and related to daily social life.
\end{itemize}

\subsubsection{\textbf{验证声明的挑战（多选）}\quad Claim verification challenges (Multiple choice)}
对声明进行事实验证往往会遇到许多挑战，挑战可以分为以下四类，标注者需要阅读事实验证的文章，判断验证声明时会遇到哪些挑战：
\begin{itemize}
\item 证据搜集：通过搜集证据，比如找相关的新闻，论文，法律法规等来验证声明。
\item 专家咨询：通过咨询专家或者相关人士，比如外交部发言人陈述，部委回复，记者采访等来验证声明。
\item 数值推理：通过数值的比较，趋势的分析来验证声明。
\item 多模态：通过除了文本外的其他证据，比如图片，视频，音频来验证声明。
\end{itemize}
Factual verification of a claim often encounters many challenges. The challenges are summarized into the following four categories. The annotator needs to read the fact-checking article to determine which challenges will be encountered in verifying the claim:
\begin{itemize}
\item Evidence Collection: Verify the claim by collecting evidence, such as finding relevant news, papers, laws and regulations, etc.
\item Expert Consultation: Verify the claim by consulting experts or related people, such as statements by the spokesperson of the Ministry of Foreign Affairs, replies from ministries and commissions, interviews with reporters, etc.
\item Numerical Reasoning: Verify the claim by numerical comparison, trend analysis, etc.
\item Multi-Modality: Verify the claim with other evidence besides articles, such as pictures, videos, and audio.
\end{itemize}
\end{CJK*}

\end{document}